\documentclass[english,ngerman,twocolumn]{article}
\usepackage{babel}
\usepackage[utf8]{inputenc}
\usepackage[big,online]{dgruyter}
\usepackage{microtype}
\usepackage[backend=bibtex,style=numeric-comp,doi=true,sorting=none]{biblatex}
\usepackage{graphicx, subfigure}
\usepackage{amssymb}
\usepackage[output-decimal-marker={.}]{siunitx}
\newcommand{\orcid}[1]{\href{https://orcid.org/#1}{\includegraphics[width=0.32cm]{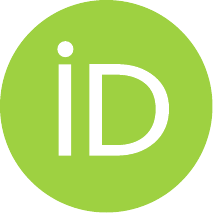}\,#1}}

\begin{document}
  \journalname{tm~–~Technisches Messen}
  \journalyear{2025}
  \journalvolume{92}
  \journalissue{S1}
  \startpage{1}

  \DOI{10.1515/teme-2025-0045  }                 

\selectlanguage{english}
\title{Enhancing 3D point accuracy of laser scanner through multi-stage convolutional neural network for applications in construction}
\runningtitle{AHMT}
\subtitle{Verbesserung der 3D-Punktgenauigkeit von Laserscannern durch ein mehrstufiges neuronales Netz für Anwendungen im Bauwesen}

\author*[1]{Qinyuan Fan}
\author[1]{Clemens Gühmann}

\affil[1]{\protect\raggedright 
 Fachgebiet Elektronische Mess- und Diagnosetechnik, Technische Universität Berlin, Einsteinufer 17, 10587 Berlin, Germany, E-mail: qinyuan.fan@tu-berlin.de \orcid{0009-0008-6521-7039}, clemens.guehmann@tu-berlin.de \orcid{0000-0002-2865-0078}}
	
\abstract{We propose a multi-stage convolutional neural network (MSCNN) based integrated method for reducing uncertainty of 3D point accuracy of lasar scanner (LS) in rough indoor rooms, providing more accurate spatial measurements for high-precision geometric model creation and renovation. Due to different equipment limitations and environmental factors, high-end and low-end LS have positional errors. Our approach pairs high-accuracy scanners (HAS) as references with corresponding low-accuracy scanners (LAS) of measurements in identical environments to quantify specific error patterns. By establishing a statistical relationship between measurement discrepancies and their spatial distribution, we develop a correction framework that combines traditional geometric processing with targeted neural network refinement. This method transforms the quantification of systematic errors into a supervised learning problem, allowing precise correction while preserving critical geometric features. Experimental results in our rough indoor rooms dataset show significant improvements in measurement accuracy, with mean square error (MSE) reductions exceeding 70\% and peak signal-to-noise ratio (PSNR) improvements of approximately 6 decibels. This approach enables low-end devices to achieve measurement uncertainty levels approaching those of high-end devices without hardware modifications.
}


\transabstract{\selectlanguage{ngerman} 
Wir schlagen eine integrierte Methode auf Basis eines mehrstufigen Convolutional Neural Network (MSCNN) vor, um die Unsicherheit der 3D-Punktgenauigkeit von Laserscanner (LS) in groben Innenräumen zu reduzieren und genauere räumliche Messungen für hochgenaue geometrische Modelle und Renovierungen zu ermöglichen. Aufgrund unterschiedlicher Gerätebeschränkungen und Umgebungsfaktoren weisen sowohl hochgenaue als auch niedriggenaue LS Positionsfehler auf. Unser Ansatz kombiniert hochgenaue Laserscanner (HAS) als Referenz mit entsprechenden Scannern mit niedriggenaue Laserscanner (LAS) für Messungen in identischen Umgebungen, um spezifische Fehlermuster zu quantifizieren. Durch die Ermittlung einer statistischen Beziehung zwischen Messabweichungen und ihrer räumlichen Verteilung entwickeln wir einen Korrekturaufbau, der traditionelle geometrische Verarbeitung mit gezielter Verfeinerung durch neuronale Netze kombiniert. Diese Methode wandelt die Quantifizierung systematischer Fehler in ein überwachtes Lernproblem um und ermöglicht so eine präzise Korrektur unter Beibehaltung kritischer geometrischer Merkmale. Experimentelle Ergebnisse in unserem Datensatz mit groben Innenräumen zeigen signifikante Verbesserungen der Messgenauigkeit mit einer Reduzierung des mittleren quadratischen Fehlers (MSE) um mehr als 70 \% und einer Verbesserung des Spitzen-Signal-Rausch-Verhältnisses (PSNR) um etwa 6 Dezibel. Dieser Ansatz ermöglicht es Low-End-Geräten, ohne Hardware-Modifikationen Messunsicherheiten zu erreichen, die denen von High-End-Geräten nahe kommen.
}

\transkeywords{Laserscanner, 3D-Punktgenauigkeit, Messunsicherheit, Convolutional Neural Network, geometrische Messung.}


\maketitle


\section{Introduction}

In recent years LS have become essential in architecture, engineering, construction (AEC) \cite{aryan_planning_2021}, and cultural heritage preservation \cite{vacca_laser_2012, haddad_ground_2011}. HAS enables digitization of environments, providing reliable data for building information modelling, structural analysis, and decision support \cite{liu_survey_2021, sanhudo_framework_2020}. Despite technological advances, laser scanning still faces limitations in cost, size, and environmental adaptability with lower-cost devices typically sacrificing 3D point accuracy, affecting applications requiring precise spatial information.

In laser scanning, technical parameters such as scanning speed, resolution, working distance, and 3D point accuracy are documented by manufacturers, with point accuracy directly affecting measurement reliability and subsequent application quality \cite{boehler_investigating_2003}. The 3D point accuracy represents measurement uncertainty \cite{reuter_numerical_2022, hageney_messunsicherheit_2025} following  metrological principles, indicating that measurements fluctuate within confidence intervals rather than representing absolute true values \cite{meyer_measurement_2007}.

3D point accuracy is affected by both internal and external factors with internal factors including measurement errors in the scanner itself, such as distance and angle measurement errors, laser beam divergence, sensor synchronization issues, geometric axis instability, and inadequate calibration \cite{zogg_investigations_2008, gerbino_influence_2016}. External factors include environmental and target conditions such as atmospheric refractive index variations, air turbulence, temperature and humidity fluctuations, dust interference, and obstacles affecting laser propagation \cite{gerbino_influence_2016}. Target characteristics are also crucial, including size, surface roughness, material reflectivity, surface curvature, laser incidence angle, and orientation, all impacting echo signal strength and measurement accuracy \cite{gerbino_influence_2016}. Additionally, scanning configuration parameters like sensor-target distance, scanning angle, and point cloud (PC) resolution determine data detail and noise levels \cite{gerbino_influence_2016}. The combined effect of these factors causes inevitable deviations in the acquired PC data from true values, manifesting as noise, offset, or distortion.

Various calibration and optimization methods have been proposed to reduce uncertainty, including hardware calibration through regular maintenance and precise parameter calibration \cite{holst_calibration_nodate}. High-precision calibration plates or targets effectively reduce instrument errors \cite{fan_high-precision_2023, jiang_accurate_2021, martinez-pellitero_new_2018, qiang_calibration_2009}. Environmental compensation focuses on correcting for atmospheric attenuation, humidity, and temperature fluctuations \cite{kaasalainen_effect_2010, reshetyuk_self-calibration_nodate, scaramuzza_extrinsic_2007, anand_evaluation_2022}. In mobile mapping scenarios, dynamic error compensation integrates inertial measurement unit data \cite{li_spatiotemporal_2022, pan_slam-based_2023, xu_fuzzy_2017}. Increasing scan density enhances PC resolution \cite{kandare_effects_2016, kellner_new_2019, tang_quantification_2009}. The application of extinction spray optimizes scanning results \cite{tang_quantification_2009, clark_using_1997}. PC filtering improves data quality by reducing noise \cite{charron_-noising_2018, duan_low-complexity_2021, gao_reflective_2022, ren_overall_2021}. Reflectance modelling optimizes echo signal processing \cite{baribeau_color_1992, he_3d_2022, hofle_correction_2007, tan_specular_2017, wang_modeling_2014, yang_dealing_2008}. Improved signal processing algorithms reduce measurement deviations \cite{pfennigbauer_improving_2010, rodriguez-quinonez_surface_2013}. Data quality assessment frameworks help quantify accuracy \cite{anand_evaluation_2022, rodriguez-quinonez_surface_2013}. While these measures improve PC quality, they cannot fully address all factors or restore true geometry. Traditional approaches often require professional expertise or additional equipment, making high-precision scanning inaccessible for budget-constrained users. Therefore, achieving accurate PC scanning at lower costs remains a significant challenge.

\begin{figure*}[htbp]
\centering
\includegraphics[width=\textwidth]{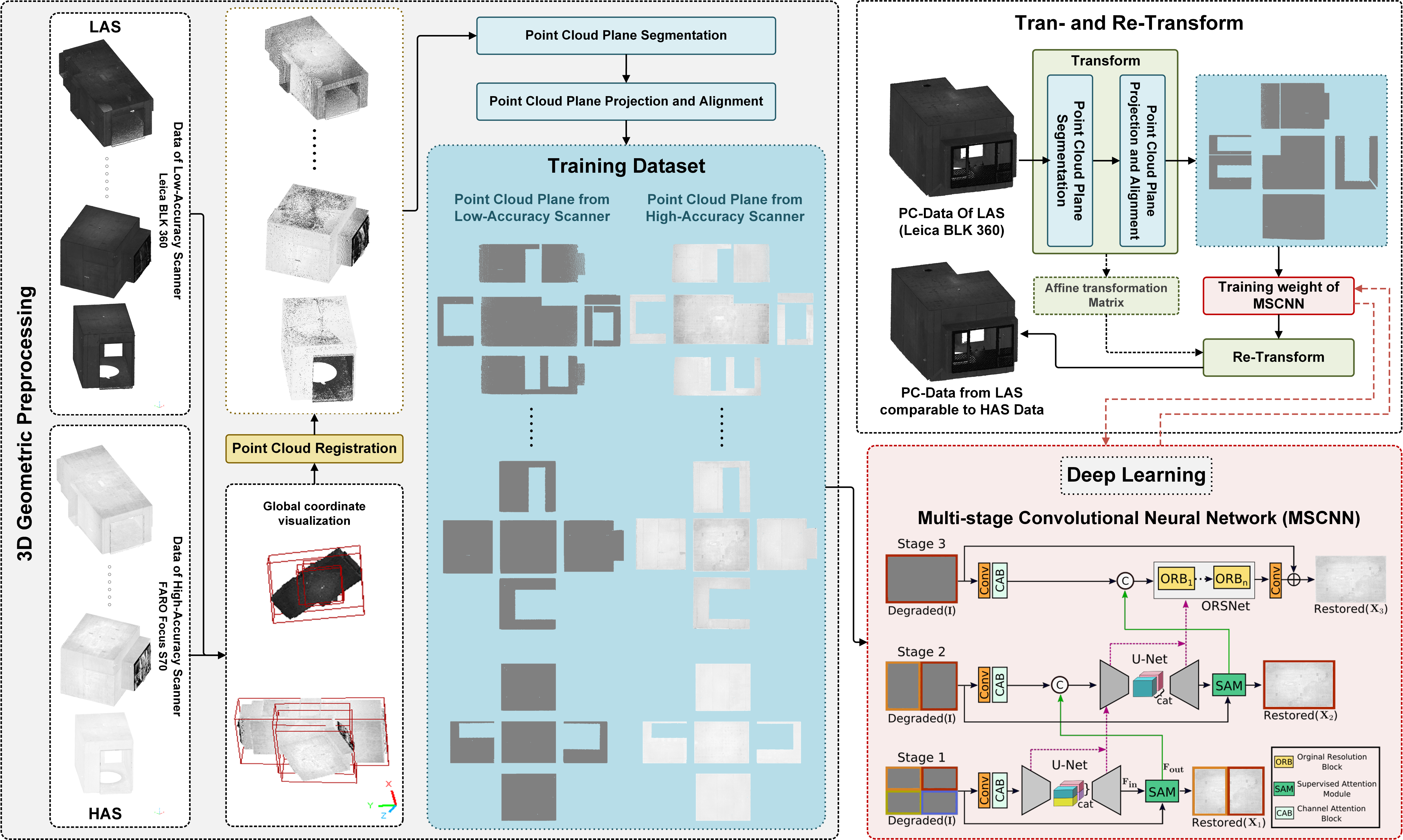}
\caption{The framework structure of the overall work}
\label{fig:AHMT_Arch}
\end{figure*}

To address measurement uncertainty challenges, we propose a 3D point accuracy optimization framework based on Multi-Scale Convolutional Neural Networks (MSCNN). Traditional methods mainly rely on physical calibration or empirical models, whereas this work introduces deep learning technology to model systematic errors as a supervised learning problem. The method uses HAS data as reference to spatially register LAS data and extract error patterns, converting 3D scene errors to 2D plane errors. An error mapping model based on MSCNN learns error distribution under different environmental and equipment conditions, achieving accurate error compensation while retaining key geometric features.

\begin{enumerate} 
\item We propose an integrated framework combining geometric processing and deep learning that reduces LAS uncertainty without hardware modifications by converting 3D scenes to 2D planes, achieving HAS-level point accuracy. 
\item MSCNN optimized for 2D surface error compensation learns multiscale features to correct common low-end scanner errors and enhance geometric quality. 
\item Real-world testing with Leica BLK360 and FARO Focus S70 demonstrated over 70\% MSE reduction and approximately 6 dB PSNR improvement, significantly enhancing measurement accuracy without hardware upgrades. 
\end{enumerate}

\section{Methods}

The overall framework comprises three integrated components, as illustrated in figure~\ref{fig:AHMT_Arch}. First, in section~\ref{3DGeometricPreprocessing}, we perform 3D geometric preprocessing to convert the room framework into a dataset of wall. Then, in section~\ref{MSCNN}, we introduce how we optimize the quality of the wall point cloud using MSCNN. Finally, section~\ref{Workflow} presents the comprehensive workflow and deployment methodology of the entire framework, including practical implementation considerations and system integration approaches.

\subsection{3D Geometric Preprocessing} \label{3DGeometricPreprocessing}
To enable accurate 3D reconstruction, we align and preprocess point clouds (PCs) from low-accuracy scans (LAS) and high-accuracy scans (HAS), denoted as $\mathcal{S}_L^{(j)}$ and $\mathcal{S}_H^{(j)}$ respectively for each scene $j \in \{1, \dots, M\}$. Each PC is a set of 3D points: $\mathcal{S}_L^{(j)} = \left\{ \mathbf{p}_i^{L(j)} \in \mathbb{R}^3 \;\middle|\; i = 1, \dots, N \right\}$, similarly for HAS.

\textbf{PC registration}: 
Since LAS and HAS devices have independent coordinate systems, coordinate deviations exist between PCs collected in the same scene, as shown in figure~\ref{fig:AHMT_Arch} (c). We first need to align the LAS $\mathcal{S}_L^{(j)}$ and HAS $\mathcal{S}_H^{(j)}$ point clouds to prepare for subsequent feature processing and all subsequent steps.

We use Fast Point Feature Histograms (FPFH) to extract local features from downsampled point clouds. For each source point $\mathbf{p}_{i}^{L(j)}$ in the LAS point cloud, we identify the most similar target point $\mathbf{p}_{l}^{H(j)}$ in the HAS point cloud by minimizing the feature distance:
\begin{equation}
\setlength{\arraycolsep}{1pt}
\mathcal{C}^{(j)} = {\scriptstyle 
\left\{ (i, l) \,\middle|\, 
l = \arg\min_{l} \left\| \mathrm{FPFH}\left(\mathbf{p}_{i}^{L(j)}\right) - \mathrm{FPFH}\left(\mathbf{p}_{l}^{H(j)}\right) \right\| 
\right\}
}
\label{eq:2}
\end{equation}
Optionally, a mutual consistency filter or a feature distance threshold can be applied to ensure robust correspondences.

We estimate a global rigid transformation using random sample consensus (RANSAC) \cite{fischler_random_1987} on the matched pairs $\mathcal{C}^{(j)}$. Given $r$ randomly sampled pairs, the optimal rotation $\mathbf{R}$ and translation $\mathbf{t}$ minimize the point-to-point Euclidean alignment error:
\begin{equation}
    \mathbf{R}^{(j)}, \mathbf{t}^{(j)} = \arg\min_{\mathbf{R}, \mathbf{t}} \sum_{k=1}^{r} \left\| \mathbf{R} \mathbf{p}_{i_k}^{L(j)} + \mathbf{t} - \mathbf{p}_{l_k}^{H(j)} \right\|^2
\end{equation}
The transformation maximizing the number of inliers within threshold $\epsilon_1 = 0.03$\,m is selected:
\begin{equation}
    \max_{\mathbf{R}, \mathbf{t}} 
    \left| 
        \left\{ (i, l) \;\middle|\; 
        \left\| \mathbf{R} \mathbf{p}_i^{L(j)} + \mathbf{t} - \mathbf{p}_l^{H(j)} \right\| < \epsilon_1 
        \right\} 
    \right|
    \label{eq:3}
\end{equation}
The resulting transformation is represented as:
\begin{equation}
\mathbf{T}^{*(j)}_{\text{global}} =
\left[
\begin{array}{cc}
\mathbf{R}^* & \mathbf{t}^* \\
\mathbf{0}^\top & 1
\end{array}
\right]
\end{equation}
To improve registration accuracy, we perform Iterative Closest Point (ICP) \cite{besl_method_1992} optimisation starting from the global transformation $\mathbf{T}^{*(j)}_{\text{global}}$ using a point-to-plane error metric:
\begin{equation}
\setlength{\arraycolsep}{1pt}
 \mathbf{T}_{\text{final}}^{(j)} = 
{\scriptstyle \arg\min_{\mathbf{R},\, \mathbf{t}} 
\sum\nolimits_{(i,l)} 
\left( \mathbf{n}_l^{H(j)} \cdot 
\left( \mathbf{R} \mathbf{p}_i^{L(j)} + \mathbf{t} - \mathbf{p}_l^{H(j)} \right)
\right)^2}
\end{equation}
where $\mathbf{n}_l^{H(j)}$ is the surface normal at $\mathbf{p}_l^{H(j)}$. Correspondences are filtered by:
\begin{equation}
    \left\| \mathbf{p}_l^{H(j)} - \left( \mathbf{R} \mathbf{p}_i^{L(j)} + \mathbf{t} \right) \right\| < \epsilon_i
\end{equation}
We adopt a three-stage coarse-to-fine refinement with decreasing thresholds $\epsilon_i \in \{0.008,\ 0.004,\ 0.002\}$ m, enabling robust initial alignment and fine-scale precision.

\textbf{PC plane segmentation}: 
After PC registration of $\mathcal{S}_L^{(j)}$  and $\mathcal{S}_H^{(j)}$, we get the set of aligned point clouds $\mathcal{S}^{(j)}$ under the unified coordinate system, and further process the fused point clouds to extract dominant planar surfaces such as  walls, floors, or ceilings, etc, as shown in figure~\ref{fig:AHMT_Arch} (d). With this step, we can convert the 3D scene into a 2D depth map, which is a preliminary preparation for subsequent ground 2D projection and flat surface alignment. We segment PC based on normal vector clustering with Density-Based Spatial Clustering of Applications with Noise (DBSCAN) density clustering to obtain a planar subset $\mathcal{P}_k^{(j)} \subset \mathcal{S}^{(j)}$.

\textbf{PC Plane Projection and Alignment}: For each segmented planar subset \(\mathcal{P}_k^{(j)} = \{ \mathbf{p}_i \in \mathbb{R}^3 \mid i = 1, \dots, N_k \}\), 
we estimate the best-fitting plane using singular value decomposition (SVD). 
We begin by computing the centroid of the point set:
\begin{equation}
\mathbf{c}_k^{(j)} = \frac{1}{N_k} \sum_{i=1}^{N_k} \mathbf{p}_i
\label{eq:centroid}
\end{equation}
Next, we subtract the centroid from each point to form a zero-mean matrix:
\begin{equation}
\mathbf{X}_k^{(j)}[i] = \mathbf{p}_i - \mathbf{c}_k^{(j)}, \quad \text{for } i = 1, \dots, N_k
\end{equation}
We apply SVD to obtain:
\begin{equation}
\mathbf{X}_k^{(j)} = \mathbf{U}_k \boldsymbol{\Sigma}_k \mathbf{V}_k^\top
\end{equation}
The last right-singular vector (i.e., the third column of \(\mathbf{V}_k\)) represents the direction of minimal variance and is selected as the unit normal vector of the plane:
\begin{equation}
\mathbf{n}_k^{(j)} = \mathbf{v}_3^{(k)}
\end{equation}
Plane fitting provides a stable local coordinate system with the planar region aligned to the \(xy\)-plane. The normal vector defines the local \(z\)-axis, and deviations along it reflect fine geometric variations from the ideal surface. These residuals are used for subsequent 2D plane-level supervision.
Using the estimated normal $\mathbf{n}_k^{(j)}$ and centroid $\mathbf{c}_k^{(j)}$, we construct a local orthonormal coordinate system where $\mathbf{n}_k^{(j)}$ defines the $z$-axis. A rigid transformation $\mathbf{T}_k \in \mathbb{R}^{4 \times 4}$ is then applied to map global 3D points to the local frame:
\begin{equation}
\tilde{\mathbf{p}}_i = \mathbf{T}_k 
\begin{bmatrix}
\mathbf{p}_i \\ 1
\end{bmatrix}
= (\tilde{x}_i, \tilde{y}_i, \tilde{z}_i)
\end{equation}
To convert 3D coordinates into discrete 2D coordinates, we apply non-uniform scaling:
\begin{equation}
\hat{\mathbf{p}}_i = (100 \tilde{x}_i,\ 100 \tilde{y}_i,\ 1000 \tilde{z}_i)
\end{equation}
Spatial axes are scaled by 100 to convert meters to centimetres, aligning pixels to a 1 cm resolution. The depth component is scaled by 1000 to convert to millimetres, enhancing local contrast and depth gradients for learning. Coordinates are then shifted to the top-left origin of the 2D plane.
\begin{equation}
\hat{\mathbf{p}}_i' = \hat{\mathbf{p}}_i - \min_j(\hat{\mathbf{p}}_j)
\end{equation}
Letting \((\hat{x}_i, \hat{y}_i)\) denote pixel locations and \(\hat{z}_i\) the projected depth, we populate a sparse depth map \(\mathbf{D}_k \in \mathbb{R}^{H \times W}\) as:
\begin{equation}
\mathbf{D}_k[\hat{y}_i, \hat{x}_i] = \hat{z}_i
\end{equation}
Unoccupied pixels are set to NaN. To normalize the depth range across different planes, we center the values as:
\begin{equation}
\hat{z}_i \leftarrow \hat{z}_i - \frac{\max(\hat{z})}{2}
\end{equation}
The resulting depth maps have a depth distribution containing both positive and negative values and are saved as \texttt{.npy} files as a training dataset for the next step, as shown in figure~\ref{fig:AHMT_Arch} (f). Through this pipeline, we created a structured data representation that solved the problem of uneven spatial distribution in point clouds, making the data more suitable for processing by convolutional neural networks. This conversion not only improved computational efficiency but also fully utilized the advantages of CNNs in image processing. In addition, this method retained key geometric features during the conversion process and transformed the problem of optimizing 3D point accuracy into a problem that can be solved through supervised learning.

\subsection{MSCNN} \label{MSCNN}
In section~\ref{3DGeometricPreprocessing}, we convert the 3D scene into a 2D plane. Building upon this, we propose a multi-stage convolutional neural network (MSCNN), which improves upon multi-stage progressive image restoration (MPRNet)~\cite{zamir_multi-stage_2021} while maintaining its fundamental architectural elements, including the multi-stage encoder-decoder structure, channel attention blocks (CAB), and supervised attention modules (SAM), as shown in the figure~\ref{fig:AHMT_Arch} (g).

We chose to improve upon MPRNet because its multi-stage processing structure and residual prediction strategy are particularly well suited to the task of correcting geometric errors in laser scans. Its channel attention and supervision attention modules effectively retain key geometric features. It differs from Transformer~\cite{vaswani_attention_2017} in that it does not require powerful hardware support, maintaining computational efficiency while delivering satisfactory performance. This makes it more suitable for large-scale scenarios in the construction.



Unlike directly regressing the restored surface at each stage, MSCNN adopts a residual prediction strategy. Specifically, at each stage $S$, the model predicts a residual $\mathbf{R}_S$ that is added to the input $\hat{\mathbf{p}}_i^{L'}$ to obtain:
\begin{equation}
\mathbf{X}_S = \hat{\mathbf{p}}_i^{L'} + \mathbf{R}_S,
\end{equation}
where $\mathbf{X}_S $ denotes the refined surface prediction at stage $S$, and $\hat{\mathbf{p}}_i^{H'}$ is the high-precision reference. This residual-based formulation simplifies the learning task by focusing the network on correcting local deviations rather than reconstructing the entire surface from scratch.

The major architectural change is the replacement of all Parametric ReLU (PReLU) activations with Sigmoid Linear Units (SiLU)~\cite{elfwing_sigmoid-weighted_2018, ramachandran_searching_2017}:
\begin{equation}
    \text{SiLU}(x) = x \sigma(x) = \frac{x}{1 + e^{-x}},
\end{equation}
since the original MPRNet was designed to process image data in the [0, 255] range, while the data range in this study is [-7, 7] (representing deviations relative to the ideal plane), SiLU can effectively handle this scenario containing important negative values. It provides smooth gradient propagation across the entire domain; can process and retain critical negative information, fully capturing wall surface concavity features; is more sensitive to subtle geometric changes; and helps mitigate the gradient vanishing problem in deep networks \cite{ramachandran_searching_2017}. 

In terms of training objectives, we retain MPRNet’s loss design. The final loss L consists of the Charbonnier Loss $\mathcal{L}_{char}$ \cite{charbonnier_two_1994}, which penalizes the difference between the prediction and ground truth, and the Edge Loss $\mathcal{L}_{edge}$ \cite{charbonnier_deterministic_1997}, which focuses on preserving high-frequency details:
\begin{equation}
\mathcal{L}_{char} = \sqrt{ \| \mathbf{X}_S - \hat{\mathbf{p}}_i^{H'} \|^2 + \varepsilon^2 }
\end{equation}
\begin{equation}
\mathcal{L}_{edge} = \sqrt{ \| \Delta(\mathbf{X}_S) - \Delta(\hat{\mathbf{p}}_i^{H'}) \|^2 + \varepsilon^2 }
\end{equation}
\begin{equation}
\mathcal{L} = \sum_{S=1}^{3} \left[ \mathcal{L}_{char}(\mathbf{X}_S, \hat{\mathbf{p}}_i^{H'}) + \lambda \mathcal{L}_{edge}(\mathbf{X}_S, \hat{\mathbf{p}}_i^{H'}) \right]
\end{equation}

By leveraging residual learning, SiLU activation, and MPRNet’s proven multi-stage design, MSCNN achieves more stable optimization, enhanced feature flexibility, and finer recovery of complex geometric surfaces \cite{ramachandran_searching_2017}. The specific results will be demonstrated in sectionn~\ref{3section}.

\subsection{Workflow: Trans- and Re-Transform} \label{Workflow}

To ensure high accuracy and robustness in deep learning models for processing LAS point cloud data, we adopt a Transform and Re-Transform workflow, as shown in figure~\ref{fig:AHMT_Arch} (h). The main objective of this workflow is to align the input data so that newly acquired 3D scenes are represented within a unified local coordinate system that matches the representation learned by the MSCNN model during training. This alignment enhances the effectiveness of the pretrained MSCNN weights.

In the Transform stage, we apply PC Plane Segmentation and PC Plane Projection and Alignment (as described in section~\ref{3DGeometricPreprocessing}) to rotate and translate each 2D surface of the 3D scene into its corresponding local planar coordinate system. During this process, affine transformation matrices are recorded for each surface. This step helps reduce geometric inconsistencies caused by variations in viewpoint, occlusion, and spatial deviation in the raw data.

After alignment, the MSCNN model is applied to each transformed 2D surface for restoration and enhancement. In the Re-Transform stage, the previously saved affine matrices are used to project the model outputs back to their original 3D spatial positions, ensuring geometric consistency. This end-to-end workflow enables global optimization of the 3D scene, significantly improves the quality of the LAS data, and indirectly enhances the 3D point accuracy of the LAS device.

\begin{figure*}[!htb]
\centering
\includegraphics[width=\textwidth, trim=6cm 0cm 0cm 1cm, clip]{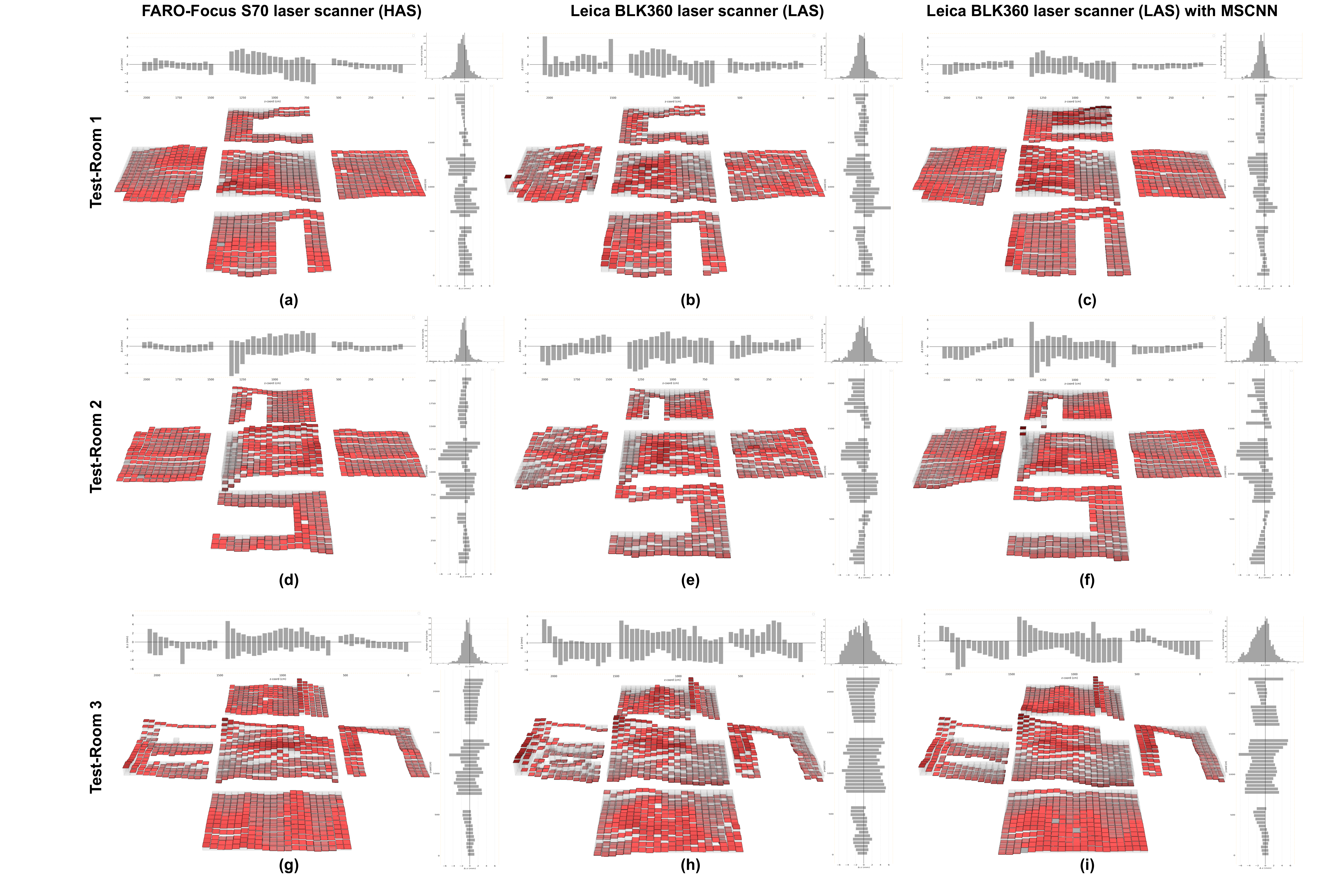}
\caption{Quality Comparison of Indoor Point Cloud Data from different rough rooms - scanning results from high-precision FARO-Focus S70, original Leica BLK360 and MSCNN-optimized Leica BLK360. The base of the bar chart is 20 x 20 cm.The lighter the red color, the better the point cloud quality; the darker the red color, the poorer the point cloud quality. Gray areas indicate nearly perfect sections.}
\label{fig:AHMT_result}
\end{figure*}

\section{Experimental Results and Discussion} \label{3section}

This section details the experimental validation of our multi-stage convolutional neural network (MSCNN) method for enhancing 3D point accuracy of laser scanners in construction applications.

\subsection{Data collection} \label{Datacollection}
For the experimental validation of our MSCNN-based approach to improve 3D point accuracy, we used two commercial laser scanners with different accuracy specifications: the FARO Focus S70 as a high-accuracy scanner (HAS) and the Leica BLK360 as a lower-accuracy scanner (LAS). Table~\ref{tab:scanner_accuracy} summarizes the key specifications of these devices.
\begin{table}[ht]
\caption{\label{tab:scanner_accuracy}Key product specifications of the laser scanners}
\begin{tabular}{cc}
\textbf{Laser Scanner} & \textbf{3D Point Accuracy} \\ \hline
FARO Focus S70 (HAS) & \SI{2}{\milli\meter} \\
Leica BLK360 (LAS) & \SI{4}{\milli\meter} \\
\end{tabular}
\end{table}
Data acquisition was conducted in 50 indoor environments with varying surface conditions. To ensure consistency, both devices were positioned at identical locations within each room, and scanning was performed under controlled conditions to minimize environmental interference. Instruments were visually centred over the station points without the use of a sighting tool nor a plumb bob. Figure~\ref{fig:AHMT_Arch} (a) and (b) show the visualisation of PC data of HAS and LAS.

\subsection{Experimental setup}
Based on the 50 indoor point clouds collected under the conditions described in section~\ref{Datacollection}, we used 40 rooms as the training set and 10 rooms as the test set. We performed 3D geometric preprocessing on the collected 50 indoor point cloud data according to the steps in section~\ref{3DGeometricPreprocessing}, obtaining 275 indoor wall data. Among them, the 40 rooms in the training set had 221 walls, and the 10 rooms in the test set had 54 walls.

\subsection{Evaluation metrics}
We evaluate the performance of our MSCNN model using MSE and PSNR, which are mathematically defined as follows:
\begin{align} 
\text{MSE} = \frac{1}{N} \sum_{i=1}^{N} \left( \hat{\mathbf{p}}_i^{H'} - \mathbf{X}_S \right)^2 
\end{align}
\begin{align} 
\text{PSNR} = 20 \cdot \log_{10}\left(\frac{\text{14}}{\sqrt{\text{MSE}}}\right) 
\end{align}
Where $\hat{\mathbf{p}}_i^{H'}$ is the high-precision reference (ground truth from HAS); $\mathbf{X}_S$ is the final output prediction from our model at stage $S$; $N$ is the number of points in the point cloud; and $14$ is the maximum possible depth difference value $(-7,7)$ in our normalized depth representation.

These metrics quantify our model's ability to enhance the accuracy of low-cost laser scanner data. MSE measures the average squared difference between the estimated surface and the high-precision reference surface, with lower values indicating smaller prediction errors. PSNR represents the ratio between the maximum possible depth value and the power of corrupting noise, with higher values indicating better quality reconstruction.

\subsection{Result analysis}

\begin{table}[ht]
\begin{center}
\caption{\small Comparison of the results of the original Leica BLK360 (LAS) data and FARO Focus S70 (HAS) data evaluation with the results of the Leica BLK 360 (LAS) data improved by the MSCNN-based integration method and the FARO Focus S70 (HAS) data evaluation.}
\label{table:result}
\vspace{-2mm}
\setlength{\tabcolsep}{3pt}  
\begin{tabular}{p{1.1cm} c | c | c || c | c }
 & & \multicolumn{2}{c||}{Original BLK360} & \multicolumn{2}{c}{Improved BLK360} \\
Scenes  & Wall count & MSE~$\textcolor{black}{\downarrow}$ & PSNR~$\textcolor{black}{\uparrow}$ & MSE~$\textcolor{black}{\downarrow}$ & PSNR~$\textcolor{black}{\uparrow}$\\
\midrule[0.15em]
Room 1  & 5 & 4.588 & 23.040 & 1.130 & 29.472 \\ 
Room 2  & 5 & 4.830 & 22.854 & 1.234 & 28.778 \\ 
Room 3  & 7 & 4.889 & 22.941 & 1.417 & 28.316 \\ 
Room 4  & 5 & 4.666 & 23.102 & 1.056 & 29.700 \\ 
Room 5  & 5 & 5.323 & 22.388 & 1.260 & 28.790 \\ 
Room 6  & 5 & 5.022 & 22.634 & 1.288 & 28.938 \\ 
Room 7  & 7 & 5.744 & 22.310 & 1.589 & 28.978 \\ 
Room 8  & 5 & 4.614 & 23.132 & 1.142 & 29.922 \\ 
Room 9  & 5 & 6.398 & 21.852 & 1.442 & 28.334 \\ 
Room 10  & 5 & 5.388 & 22.650 & 1.358 & 29.342 \\ 
\bottomrule[0.1em]
\textbf{All} & 54 & \textbf{5.145} & \textbf{22.690} & \textbf{1.292} & \textbf{29.057} \\
\end{tabular}
\end{center}\vspace{-1.5em}
\end{table}

The experimental results show in table~\ref{table:result} significant improvements when using our MSCNN based integrated method to enhance low-accuracy laser scanner data. Figure~\ref{fig:AHMT_result} displays three test rooms, comparing point clouds from the FARO-Focus S70 (HAS), original Leica BLK360 (LAS), and BLK360 with our MSCNN. The original BLK360 scans are shown in figure~\ref{fig:AHMT_result} (b), (e), (h) show noticeable noise and irregularities across all test rooms, with scattered points and uneven surfaces reflecting the systematic errors in lower-cost hardware. In contrast, the MSCNN based integrated method corrected scans are shown in figure~\ref{fig:AHMT_result} (c), (f), (i) display smoother surfaces and more consistent point distributions that closely match the high-accuracy reference data. Histograms in the upper corners confirm these improvements, with our correction producing narrower, more peaked distributions similar to the FARO-Focus S70 reference shown in figure~\ref{fig:AHMT_result} (a), (d), (g), indicating reduced measurement variance.

In addition to the qualitative assessment results presented in figure~\ref{fig:AHMT_result}, our quantitative analysis in table~\ref{table:result} provides compelling statistical evidence of our method's effectiveness. Our approach achieved a 74.9\% reduction in average MSE (from 5.145 to 1.292) and enhanced PSNR by 6.367 dB (from 22.690 to 29.057) across all 54 walls from the 10 test rooms. Room 9 exhibited the most dramatic improvement with a 77.5\% MSE reduction, while Room 4 achieved the best overall performance (MSE: 1.056, PSNR: 29.700 dB). Notably, even architecturally complex environments such as Rooms 3 and 7 (each containing 7 walls) showed substantial gains, confirming our method's robustness across diverse spatial configurations. These quantitative improvements align with the visual quality enhancements demonstrated in figure~\ref{fig:AHMT_result}, where the reduction in artifacts and improved structural coherence can be clearly observed across various room types.

\section{Conclusion}
In this paper, an integrated method based on multi-stage convolutional neural network (MSCNN) is proposed to improve the 3D point accuracy of a low-precision laser scanner by the real data from a high-precision laser scanner in an indoor scene. The uncertainty of 3D point accuracy in indoor laser scanning environments is successfully reduced by pairing the measurements of high-precision and low-precision scanners in the same environment to quantify the error patterns. The correction framework we developed transforms the systematic error quantization into a supervised learning problem, which achieves accurate correction while preserving key geometric features. Experimental results show significant improvement in our method, which exhibits remarkable robustness even in complex indoor environments. Future research directions include extending the dataset to cover more diverse environments, expanding to outdoor application scenarios and high-precision industrial scenarios, optimizing the 3D point accuracy of scanning devices dedicated to different tasks, improving the accuracy of geometrical measurements in the industrial field, enhancing the accuracy of downstream tasks such as CAD modelling, realizing real-time processing capabilities, and exploring cross-device generalization. We expect that this research will provide new perspectives for laser scanning to improve measurement accuracy, measurement uncertainty, and development of more accurate measurement systems, and at the same time provide higher-quality point cloud data inputs for downstream applications, such as CAD modeling, and promote the innovative application of deep learning technology in 3D geometric measurement.

\begin{acknowledgement}

The work presented was financed by the Technische Universität Berlin. Our data comes from a newly constructed real estate development. The entire data collection process was completed by Mr. Mingchang Chen. We would like to express our sincere gratitude to Mr. Chen for his valuable contributions to this project.

\end{acknowledgement}

\printbibliography

\end{document}